\documentclass[twocolumn,times,final,authoryear]{elsarticle}
\usepackage{graphicx} 
\usepackage{amsmath}
\usepackage{amssymb}
\usepackage[dvipsnames,svgnames]{xcolor}
\usepackage[T1]{fontenc}
\usepackage{framed}
\usepackage{multirow}
\usepackage{dirtytalk}
\usepackage{tikz}
\usepackage{pgfplots}
\usepackage{booktabs}
\usepackage{natbib}
\pgfplotsset{compat=newest}
\usetikzlibrary{external}
\usepgfplotslibrary{groupplots}

\newcommand{\parsection}[1]{\textbf{#1}}

\journal{Pattern Recognition Letters \\ https://doi.org/10.1016/j.patrec.2024.09.011}

\DeclareMathOperator{\MAE}{MAE}
\renewcommand{\AE}{\operatorname{AE}}
\DeclareMathOperator{\AUSE}{AUSE}

\newcommand{\x}{\mathbf{x}}
\newcommand{\E}{\mathcal{E}}

\definecolor{deep1}{RGB}{76, 114, 176}
\definecolor{deep2}{RGB}{221, 132, 82}
\definecolor{deep3}{RGB}{85, 168, 104}
\definecolor{deep4}{RGB}{196, 78, 82}
\definecolor{deep5}{RGB}{129, 114, 179}
\definecolor{deep6}{RGB}{147, 120, 96}
\definecolor{deep7}{RGB}{218, 139, 195}
\definecolor{deep8}{RGB}{140, 140, 140}
\definecolor{deep9}{RGB}{204, 185, 116}
\definecolor{deep10}{RGB}{100, 181, 205}

\pgfplotsset{
    cycle list={
        deep1,every mark/.append style={fill=deep1},mark=*\\
        deep2,every mark/.append style={fill=deep2},mark=square*\\
        deep3,every mark/.append style={fill=deep3},mark=otimes*\\
        deep4,mark=star\\
        deep5,every mark/.append style={fill=deep5},mark=diamond*\\
        deep6,densely dashed,every mark/.append style={solid,fill=deep6},mark=*\\
        deep7,densely dashed, every mark/.append style={solid,fill=deep7},mark=square*\\
        deep8,every mark/.append style={solid,fill=deep8},mark=otimes*\\
        deep9,densely dashed,mark=star,every mark/.append style=solid\\
        deep10,densely dashed,every mark/.append style={solid,fill=deep10},mark=diamond*\\
    },
    /tikz/mark size=1pt
}

\begin{document}

\begin{frontmatter}

\title{Uncertainty Quantification Metrics for Deep regression}

\author[1]{Simon Kristoffersson Lind\corref{cor1}\fnref{fn1}}
\cortext[cor1]{Corresponding author:}
\ead{simon.kristoffersson\_lind@cs.lth.se}

\author[2]{Ziliang Xiong\fnref{fn1}} 
\ead{ziliang.xiong@liu.se}
\author[2]{Per-Erik Forss\'en}
\ead{per-erik.forssen@liu.se}
\author[1]{Volker Kr\"uger}
\ead{volker.krueger@cs.lth.se}

\fntext[fn1]{Equal contribution}


\affiliation[1]{
    organization={Lund University LTH},
    city={Lund}, 
    country={Sweden},
}

\affiliation[2]{
    organization={Linköping University},
    city={Linköping}, 
    country={Sweden},
}


\begin{abstract}
When deploying deep neural networks
on robots or other physical systems,
the learned model should reliably quantify predictive uncertainty. A reliable uncertainty allows 
downstream modules to reason about the safety 
of its actions.
In this work, we address metrics for uncertainty quantification. 
Specifically, we focus on regression tasks, and investigate Area Under Sparsification Error (AUSE),
Calibration Error (CE), Spearman's Rank Correlation, and Negative Log-Likelihood (NLL).
Using multiple datasets, we look into how those metrics behave under four typical types of uncertainty, their stability regarding the size of the test set, and reveal their strengths and weaknesses.
Our results indicate that Calibration Error is the most stable and interpretable metric, but AUSE and NLL also have their respective use cases.
We discourage the usage of Spearman's Rank Correlation for evaluating uncertainties 
and recommend replacing it with AUSE.
\end{abstract}



\end{frontmatter}

\section{Introduction}
In recent years, there has been a rapid advance in the adoption of neural network-based methods in many real-world applications, for example robotics.
Following this adoption,
increasing scrutiny has been directed towards neural network-based methods for their lack of reliability and interpretability.
While neural networks have achieved impressive performance for many different tasks,
the fact remains that they can be unreliable in real-world deployment \citep{grimmett2016introspection}.
Additionally, their lack of interpretability makes it difficult to know how and when they may perform unreliably.
For these reasons, increasing attention has been directed at the uncertainty output from neural networks,
and the importance of introspective qualities \citep{grimmett2016introspection}.
Arguably the most important introspective quality is a reliable uncertainty estimate.

\begin{figure}[t]
        \centering
        \includegraphics[width=\columnwidth]{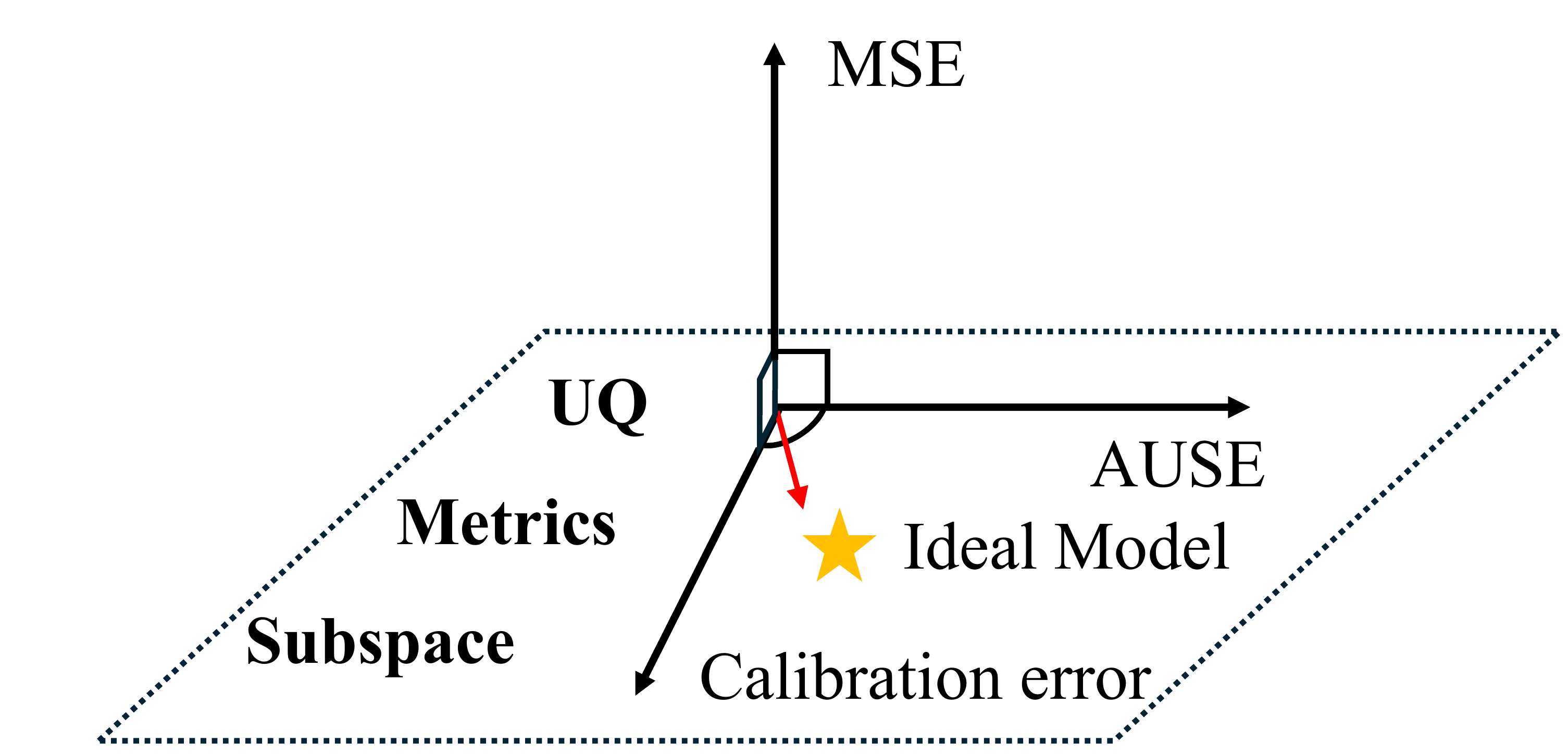} 
	\caption{An illustration for UQ metrics and regression metrics. Note: the axes of CE and AUSE are distinct, but not orthogonal.}
 \label{fig:illustration}  
\end{figure}

Despite increasing attention being directed toward \textit{uncertainty quantification} (UQ), much of this work focuses on uncertainty in classification tasks where metrics for assessment are well understood \citep{ovadia2019can}.
Many real-world applications, however, rely instead on regression,
and there is a lack of common understanding surrounding the available metrics for regression.
In this work, we have identified four metrics that are commonly used to measure various qualities in a predicted uncertainty in regression.
Specifically, we investigate the {\it Area Under Sparsification Error} (AUSE) \citep{ilg2018uncertainty},
{\it Spearman's Rank Correlation} \citep{spearman1904}, {\it Negative Log Likelihood} (NLL)\citep{lakshminarayanan2017simple},
and \textit{Calibration Error} (CE) \citep{naeini2015}. 
These UQ metrics measure different aspects of uncertainty that are all orthogonal to the regression task performance (typically {\it mean squared error} (MSE)) as indicated in Fig.~\ref{fig:illustration}.

With the help of synthetic datasets, we explore the usage of these metrics with the aim of better understanding their differences, strengths and limitations.

Our contributions are as follows:
\vspace{-1ex}
\begin{itemize}
    \itemsep-0.5ex
    \item We create simple synthetic datasets that highlight different types of uncertainty, and use these to explore how the metrics behave, compared to the generating data distribution. 
    \item We compare each metric in terms of their stability to different dataset sizes.
    \item We reason about the strengths and limitations of each metric.
    \item Finally, we experiment on a real stereo disparity task, where we add synthetic noise to the input, to further illustrate the behaviour of the tested measures.
\end{itemize}
\vspace{-1ex}
We also provide a mathematical formulation for the AUSE metric.
To the best of our knowledge, AUSE has previously only been described informally, using natural language.

\section{Related Work}
Historically, methods that deal with any type of uncertainty in regression have been formulated using parameterized distributions.
As a classic example, consider Kalman filters \citep{kalman1960},
which are built on a parameterized distribution with well-defined uncertainty.
Modern machine learning methods are built upon a solid statistical foundation and most methods output
some form of uncertainty \citep{bishop_2006}.
However, it has been observed that there is a discrepancy between the predicted uncertainty from modern neural networks,
and the observed empirical accuracy \citep{guo2017calibration}.
Additionally, lacking interpretability in these uncertainties have led to the development of methods to formulate
more statistically grounded uncertainties \citep{gawlikowski_survey_2021}.

Along with the research on uncertainty estimation, a number of methods have been developed to assess how trustworthy a predicted uncertainty is.
Perhaps the most commonly used (for example in \citet{guo2017calibration, kuleshov2018})
is the {\it expected calibration error} (ECE) \citep{naeini2015}.
While not as common as ECE, NLL ({\it negative log likelihood}) is also used, for example in \citet{nll_ex1, nll_ex2}.
Spearman's rank correlation coefficient \citep{spearman1904} has also found new use in this area
(for example in \citet{spearman_ex1, spearman_ex2}).
Finally, the most recent method we examine is AUSE \citep{ilg2018uncertainty},
used in for example \citet{ause_ex1, xiong2023hinge}.
While other uncertainty assessment methods exist, we have chosen to focus on these four as they are most commonly used.

\section{Theory}
First, we introduce different types of uncertainty in Sec.~\ref{sec:uncetainty_type}. 
Then, in Sec.~\ref{sec:metrics}, we introduce the four common uncertainty evaluation metrics to be analyzed.
\subsection{Different types of Uncertainty}
\label{sec:uncetainty_type}
In the realm of predictive uncertainty estimation, we encounter two fundamental types of uncertainty \citep{kendall2017uncertainties}. 
The first, known as {\it aleatoric uncertainty}, encompasses the intrinsic noise and ambiguity present in observations.
This noise, stemming from sources like sensor or motion irregularities, persists even with an increase in data collection and cannot be mitigated.
The second type, \textit{epistemic uncertainty} constitutes the uncertainty surrounding model parameters, reflecting our lack of knowledge about the precise model generating the observed data. 
This form of uncertainty tends to diminish as more data is acquired, and is thus often termed {\it model uncertainty}.
The most common cause of epistemic uncertainty is {\it out-of-distribution} (OOD) data.
In other words, data that comes from a distribution that is different from the training data.
For example, autonomous driving models that are trained on synthetic data usually face domain gap to real-world data.

Aleatoric uncertainty can be subdivided into \textit{homoscedastic uncertainty}, which remains constant across various inputs, and \textit{heteroscedastic uncertainty}, which varies depending on the model inputs.
Heteroscedastic uncertainty plays a crucial role in computer vision tasks.
For example, in depth regression, images with intricate textures and prominent vanishing lines typically yield confident predictions, whereas images of featureless surfaces are expected to produce higher uncertainty.

\subsection{Uncertainty Evaluation Metrics}
\label{sec:metrics}
Here, we introduce the four selected uncertainty metrics for the deep regression tasks.
We define the metrics as functions of a dataset $S = \{(\x_i, y_i) \ | \ i=1, 2, \ldots, N\}$, consisting of input-output pairs $(\x_i, y_i)$.
For any learned model, we further define
$    F(\x; \theta) \mapsto \mathbb{R}, \ \ \text{and} \ \ U(\x; \theta) \mapsto \mathbb{R} \enskip $
to be functions based on the model parameters $\theta$.
$F(\x; \theta)$ produces a prediction of $y$, and $U(\x; \theta)$ produces an uncertainty estimate.

\parsection{AUSE}:
\textit{Sparsification plots} are widely used in optical flow \citep{6261321,8237395} and stereo disparity \citep{9561617} to
assess how well the predicted uncertainty coincides with the prediction error on a test dataset.
A sparsification plot is created by plotting the mean absolute error for a dataset while iteratively removing the data point with the highest predicted uncertainty.
If the predicted uncertainty coincides well with the error,
the sparsification plot should form a monotonically decreasing curve.
The best possible sparsification plot, commonly called \textit{oracle sparsification},
is generated when the predicted uncertainty corresponds exactly to the absolute error.
\citet{ilg2018uncertainty} define \textit{sparsification error} as the difference between a models' sparsification and the corresponding oracle.
AUSE is then defined as the area under the sparsification error curve.
However, \citet{ilg2018uncertainty} never provide a formal mathematical definition for AUSE,
and therefore we provide one here.

AUSE is computed as an aggregate metric over a dataset $S$ consisting of input-output $(\x_i,y_i)$ pairs.
We define the \textit{mean absolute error} (MAE) on $S$ as:
\begin{equation}
    \MAE(S) = \frac{1}{|S|} \sum_{(\x_i, y_i) \in S} |y_i - F(\x_i; \theta)| \enskip .
\end{equation}
Based on a parameter $\alpha\in[0,1]$, we partition $S$ into disjoint subsets $S^{\AE}_{\land}(\alpha)$ and $S^{\AE}_{\lor}(\alpha)$ such that $|S^{\AE}_{\land}(\alpha)|=\alpha|S|$, and
\begin{gather}
\begin{split}
    |y_i - F(\x_i; \theta)| &\ge |y_j - F(\x_j; \theta)| \\
    \forall \enskip (\x_i, y_i) \in S^{\AE}_{\land}(\alpha) &, \enskip (\x_j, y_j) \in S^{\AE}_{\lor}(\alpha) \enskip .
\end{split}
\end{gather}
In other words, the size of $S^{\AE}_{\land}(\alpha)$ is a fraction $\alpha$ of the entire set $S$, and all absolute errors in $S^{\AE}_{\land}(\alpha)$ are larger than those in $S^{\AE}_{\lor}(\alpha)$.
We also define analogous partitions $S_{\land}^U(\alpha)$ and $S_{\lor}^U(\alpha)$ with respect to the uncertainty:
\begin{gather}
\label{eq:uncertainty}
\begin{split}
    U(\x_i; \theta) &\ge U(\x_j; \theta) \\
    \forall \enskip (\x_i, y_i) \in S^U_{\land}(\alpha) &, \enskip (\x_j, y_j) \in S^U_{\lor}(\alpha) \enskip .
\end{split}
\end{gather}

With these partitions in place, we define:
\begin{gather}
\label{eq:ause}
    \AUSE(S)=
    \int_0^1 \frac{\MAE(S^U_{\lor}(\alpha))}{\MAE(S)}
    - \frac{\MAE(S^{\AE}_{\lor}(\alpha))}{\MAE(S)} \enskip  d\alpha .
\end{gather}
AUSE computes the normalized area between the two curves $\MAE(S_{\lor}^U(\alpha))$ and $\MAE(S_{\lor}^{\AE}(\alpha))$
formed by varying $\alpha$ from 0 to 1.
The curve $\MAE(S_{\lor}^{\AE}(\alpha))$ is called the \textit{oracle},
which represents a lower bound on sparsification. 
In practice, the integral \eqref{eq:ause} is replaced with summing over finite test set samples.
See Fig.\ \ref{fig:sparsification} for an example of an AUSE sparsification plot.

Since AUSE is computed as the difference between two curve, the definitive lower bound is zero,
but there is no clear upper bound.

\parsection{Calibration Error}
The reliability diagram \citep{niculescu2005} and the expected calibration error \citep{naeini2015} are originally diagnostic tools for classification models that compare sample accuracy against predicted confidence. 
However, this approach does not apply to regression tasks.
Therefore, \citet{kuleshov2018} introduce a calibration plot for regression tasks in terms of the cumulative distribution function from the model, and summarize it with \textit{CE} as a numerical score.
Formally, let the predicted probability distribution of an input $\x_i$ be 
$
P(y_i | \theta) = F(\x_i;\theta).
$
The empirical frequency is then defined as 
\begin{equation}
    \hat{p}_j=\frac{|\{y_i \mid P(y < y_i | \theta) \leq p_j, (\x_i, y_i)\in S\}|}{N} \enskip .
\end{equation}
Here $p_j\in [0,1]$ represents an arbitrary threshold value.
Given $M$ distinct thresholds $p_1, p_2, \ldots, p_M$ (typically evenly spaced from 0 to 1), the calibration error is defined for regression as
\begin{equation}
    \operatorname{cal}(\hat{p}_1, \cdots, \hat{p}_N) = \sum_{j=1}^M w_j (p_j-\hat{p}_j)^2 \enskip ,
\end{equation}
where $w_j$ are arbitrary scaling weights, usually $w_j = \frac{1}{N/\hat{p}_j}$.

\parsection{Spearman correlation}
In 1904, psychologist Charles Spearman defined what he called the \textit{rank method} of correlation \citep{spearman1904}.
He observed that there may exist correlations that would not be adequately captured by simple linear correlation.
Instead, he argued that more complex correlations may be adequately captured by comparing \textit{ranks} of elements.

Given a set of samples, the rank of a sample is simply the index that number would have in a list.
More formally, let $L = [u_1, u_2, \ldots, u_N]$ be a sequence of samples, then the rank of a sample is defined as:
\begin{equation}
    r(u_i) = 1 + |\{u_j \ | \ u_j < u_i, \ u_j \in L\}| \enskip .
\end{equation}
We then define the rank operation for a sequence:
\begin{equation}
    R(L) = [r(u_1), r(u_2), \ldots, r(u_N)] \enskip .
\end{equation}
Spearman's rank method is then the correlation between two rank sequences:
\begin{equation}
    \rho_{R(L_1), R(L_2)}=\frac{\operatorname{cov}(R(L_1), R(L_2))}{\sigma_{R(L_1)} \sigma_{R(L_2)}} \enskip .
\end{equation}
When used as a metric for uncertainty, this correlation is computed between the predictive uncertainty, and the absolute prediction errors \citep{spearman_ex1}.

\parsection{NLL}
The popular negative log-likelihood is proved to be a \textit{strictly proper scoring rule} \citep{lakshminarayanan2017simple}.
Given a dataset $S = \{(\x_i, y_i) \ | \ i=1, 2, \ldots, N\}$ and
a probability density function $p(y | \x)$ parameterised by learnable parameters $\theta$,
the NLL is defined as
\begin{equation}
    \operatorname{NLL}(S) = -\sum_{i=1}^N \log p(y_i | \x_i; \theta) \enskip .
\end{equation}
In expectation,
the NLL is minimized if and only if $p(y_i \ | \ \x_i; \theta)$ is equal to the true underlying data distribution \citep{hastie2009}.
As such, the NLL can also be used as a metric for uncertainty predictions,
since a model with a lower NLL does a better job (in expectation) of fitting the true data distribution.

\subsection{Regression Models with Uncertainty Predictions}
There exist many different model architectures that incorporate uncertainty predictions.
Note that our goal is not to investigate properties in the models.
We choose to only use two different models, namely an ensemble and an energy-based model.
While there are many other popular models for UQ (e.g.\ the recently introduced SNGP by \cite{liu2023SNGP}) the purpose of this paper is to evaluate the metrics, not the models.

\subsubsection{Deep Ensemble (DE)}
\label{sec:ensemble}
Ensemble learning combines the predictions from multiple individual models to achieve better predictive performance than any single model.
For estimating predictive uncertainty, {\it Deep Ensemble} (DE) \citep{lakshminarayanan2017simple} is a simple yet effective method that trains multiple models in the same architecture with different random initialization and data shuffling.
The ensemble is treated as a uniformly weighted mixture model.
In practice, for the regression task, each individual model outputs two scalars, which are interpreted as the mean and variance of a Gaussian distribution, and then it is trained to minimize the Gaussian NLL loss as in (\ref{eq:gaussian_nll}) on the training set.
\begin{equation}
\label{eq:gaussian_nll}
    -\log p_\theta\left(y_n \mid \mathbf{x}_n\right)=\frac{\log \sigma_\theta^2(\mathbf{x})}{2}+\frac{\left(y-\mu_\theta(\mathbf{x})\right)^2}{2 \sigma_\theta^2(\mathbf{x})}+\text { C }
\end{equation}
We follow the suggestion by \citet{lakshminarayanan2017simple} to train 5 models and thus get a mixture of Gaussians. 
This mixture is further approximated by a Gaussian whose mean and variance are the mean and variance of the mixture respectively.

\subsubsection{Energy Based Regression (EBR)}
Energy-based learning involves learning an energy function $\E(x)$.
The goal when learning $\E(x)$ is to assign low energy to observed samples \citep{gustafsson_arxiv19}.
Learning $\E(x)$ is in many ways analogous to learning a probability density function,
and as such, it has commonly been used for unsupervised learning tasks \citep{gustafsson_arxiv19}.
\citet{gustafsson_arxiv19} construct a supervised regression model
based on an energy function.
First, they define $f_{\theta}(x, y) \mapsto \mathbb{R}$ to be their learned energy function.
Then, they construct a probability density function:

\begin{equation}
     p(y | x; \theta) = \frac{e^{f(x, y; \theta)}}{Z(x; \theta)}, \enskip
    Z(x; \theta) = \int e^{f(x, \tilde y; \theta)} d \tilde y \enskip .
\end{equation}
Parameters $\theta$ are trained by minimizing the NLL with respect to the training data:

\begin{equation}
    \mathcal{L} = -\frac{1}{N} \sum_{i=1}^N \log Z(x_i; \theta) - f(x_i, y_i; \theta) \enskip .   
\end{equation}
$Z(x_i; \theta)$ is approximated by Monte Carlo sampling.
Predictions are generated by performing multi-start gradient ascent to maximize $f(x_i, \tilde y; \theta)$ with respect to $\tilde y$.
For uncertainty quantification, the raw value of the energy function is used.

\section{Experiments and Results}
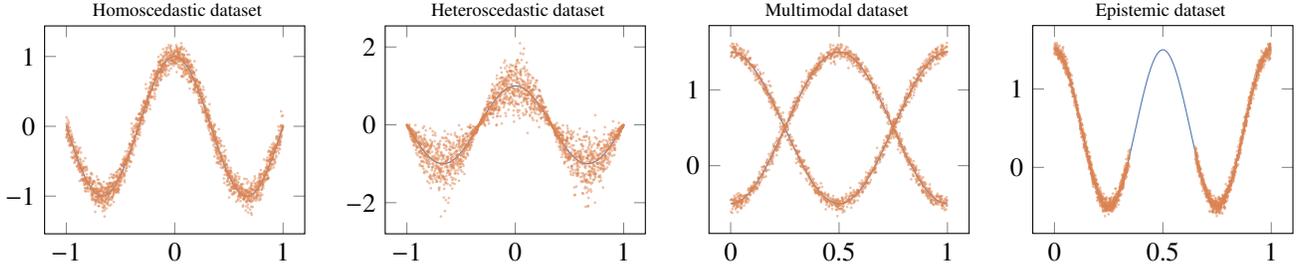
\begin{figure*}[ht!]
    \vspace{2mm}
    \begin{tabular}{@{}cccc@{}}
    \hbox{\qquad}\scriptsize{Homoscedastic dataset} & \hbox{\qquad}\scriptsize{Heteroscedastic dataset} & \hbox{\quad}\scriptsize{Multimodal dataset} & \hbox{\quad}\scriptsize{Epistemic dataset} \\
    \begin{tikzpicture}
        \begin{axis}[width=0.24\paperwidth,ylabel={y},xlabel={x}]
        \addplot [only marks, mark=*, mark size=0.3pt, opacity=0.5, color=deep2]
            table [x=train_x, y=train_y, col sep=comma] {plot_csvs/homosced_dataset.csv};
        \addplot [color=deep1, no markers, samples=100, domain=-1:1] gnuplot {cos(1.5*3.14159*x)};
        \end{axis}
    \end{tikzpicture} &
    \begin{tikzpicture}
        \begin{axis}[width=0.24\paperwidth,xlabel={x}]
        \addplot [only marks, mark=*, mark size=0.3pt, opacity=0.5, color=deep2]
            table [x=train_x, y=train_y, col sep=comma] {plot_csvs/heterosced_dataset.csv};
        \addplot [color=deep1, no markers, samples=100, domain=-1:1] gnuplot {cos(1.5*3.14159*x)};
        \end{axis}
    \end{tikzpicture} &
    \begin{tikzpicture}
        \begin{axis}[width=0.24\paperwidth,xlabel={x}]
        \addplot [only marks, mark=*, mark size=0.3pt, opacity=0.5, color=deep2]
            table [x=train_x, y=train_y, col sep=comma] {plot_csvs/multimodal_dataset.csv};
        \addplot [color=deep1, no markers, samples=100, domain=0:1] gnuplot {0.5+cos(2*3.14159*x)};
        \addplot [color=deep1, no markers, samples=100, domain=0:1] gnuplot {0.5-cos(2*3.14159*x)};
        \end{axis}
    \end{tikzpicture} &
    \begin{tikzpicture}
        \begin{axis}[width=0.24\paperwidth,xlabel={x}]
        \addplot [only marks, mark=*, mark size=0.3pt, opacity=0.5, color=deep2]
            table [x=train_x, y=train_y, col sep=comma] {plot_csvs/epistemic_dataset.csv};
        \addplot [color=deep1, no markers, samples=100, domain=0:1] gnuplot {0.5 + cos(4*3.14159*x)};
        \end{axis}
    \end{tikzpicture}
    \end{tabular}
    \vspace{-2mm}
    \caption{The four synthetic regression datasets.
        Data points are orange, and the solid blue lines represent the expectation of the generating function.}
    \label{fig:synthetic_datasets}
    \vspace{-2mm}
\end{figure*}
\subsection{Synthetic Regression Datasets}
\label{sec:datasets}
In order to gauge the behavior of the different uncertainty metrics,
we construct four simple synthetic regression datasets, each with a different source of uncertainty.
These can be seen in Fig. \ref{fig:synthetic_datasets}.
We will henceforth refer to these datasets by their names:
\textit{homoscedastic, heteroscedastic, multimodal}, and \textit{epistemic}.
The names homoscedastic and heteroscedastic are in reference to the type of Gaussian noise applied to the generating function.
Multimodal refers to the fact that data is generated from two separate generating functions.
Finally, epistemic, refers to the epistemic uncertainty arising from a gap in the training data.
Fundamentally, each dataset is a collection of input-output pairs $(x, y)$.
Outputs $y$ are generated as follows:
\begin{itemize}
    \itemsep-1ex
    \item Homoscedastic: \\[-3ex]
        $$y = \cos(1.5 \pi x) + \epsilon, \enskip \epsilon \sim \mathcal{N}(0, 0.1) \enskip .$$
    \item Heteroscedastic: \\[-3ex]
        $$y = \cos(1.5 \pi x) + \epsilon, \enskip \epsilon \sim \mathcal{N}(0, 0.4 \cdot |\cos(1.5 \pi x)|) \enskip .$$
    \item Multimodal: \\[-3ex]
        $$y = 0.5 \pm \cos(2 \pi x) + \epsilon, \enskip \epsilon \sim \mathcal{N}(0, 0.05) \enskip .$$
    \item Epistemic: \\[-3ex]
        $$y = 0.5 + \cos(4 \pi x) + \epsilon, \enskip \epsilon \sim \mathcal{N}(0, 0.05) \enskip .$$
\end{itemize}
Here, $\mathcal{N}(0, \sigma)$ refers to a Gaussian distribution with mean 0, and standard deviation $\sigma$.
$\pm$ in the definition of the multimodal dataset refers to a equal chance of being + or -, which represents the two different modes.
In the epistemic dataset, there is a gap in the training data for $x \in [0.35, 0.65]$, which is not present in the test data. This means test data is OOD, which causes epistemic uncertainty.
Domains for the inputs are $x \in [-1, 1]$ for the homoscedastic and heteroscedastic datasets,
and $x \in [0, 1]$ for the multimodal and epistemic datasets.

\subsection{Implementation Details}
\begin{figure*}[t!]
    \centering
    \vspace{2mm}
    \begin{tabular}{@{}c@{}c@{}c@{}c@{}c@{}}
    \hbox{\quad}  & \hbox{\quad}  \scriptsize{(a) Homoscedastic dataset} & \hbox{\quad\;} \scriptsize{(b) Heteroscedastic dataset} & \hbox{\quad} \scriptsize{(c) Multimodal dataset} & \hbox{\quad}\scriptsize{(d) Epistemic dataset} \\
    \makebox[5ex]{\raisebox{12ex}{\rotatebox[origin=c]{90}{\bf Deep Ensemble}}} &
    \begin{tikzpicture}
    \begin{axis}[
        width=0.24\paperwidth,
        ylabel={y},
        colormap name=viridis,
        view={0}{90},
    ]
    \addplot3 [surf,shader=interp] table [x=x, y=y, z=z, col sep=comma] {plot_csvs/ensemble_homo_contours.csv};
    \addplot3 [contour gnuplot={contour dir=z, draw color=black, number=10, labels=false}, opacity=0.2, shader=interp]
        table [x=x, y=y, z=z, col sep=comma] {plot_csvs/ensemble_homo_contours.csv};
    \addplot [only marks, mark=*, color=deep2]
        table [x=test_x, y=test_y, col sep=comma] {plot_csvs/ensemble_homo_predscatter.csv};
    \addplot [only marks, mark=*, color=deep1]
        table [x=test_x, y=pred, col sep=comma] {plot_csvs/ensemble_homo_predscatter.csv};
    \end{axis}
    \end{tikzpicture} &
    \begin{tikzpicture}
    \begin{axis}[
        width=0.24\paperwidth,
        colormap name=viridis,
        view={0}{90},
    ]
    \addplot3 [surf,shader=interp] table [x=x, y=y, z=z, col sep=comma] {plot_csvs/ensemble_hetero_contours.csv};
    \addplot3 [contour gnuplot={contour dir=z, draw color=black, number=10, labels=false}, opacity=0.2, shader=interp]
        table [x=x, y=y, z=z, col sep=comma] {plot_csvs/ensemble_hetero_contours.csv};
    \addplot [only marks, mark=*, color=deep2]
        table [x=test_x, y=test_y, col sep=comma] {plot_csvs/ensemble_hetero_predscatter.csv};
    \addplot [only marks, mark=*, color=deep1]
        table [x=test_x, y=pred, col sep=comma] {plot_csvs/ensemble_hetero_predscatter.csv};
    \end{axis}
    \end{tikzpicture} &
    \begin{tikzpicture}
    \begin{axis}[
        width=0.24\paperwidth,
        colormap name=viridis,
        view={0}{90},
    ]
    \addplot3 [surf,shader=interp] table [x=x, y=y, z=z, col sep=comma] {plot_csvs/ensemble_multimodal_contours.csv};
    \addplot3 [contour gnuplot={contour dir=z, draw color=black, number=10, labels=false}, opacity=0.2, shader=interp]
        table [x=x, y=y, z=z, col sep=comma] {plot_csvs/ensemble_multimodal_contours.csv};
    \addplot [only marks, mark=*, color=deep2]
        table [x=test_x, y=test_y, col sep=comma] {plot_csvs/ensemble_multimodal_predscatter.csv};
    \addplot [only marks, mark=*, color=deep1]
        table [x=test_x, y=pred, col sep=comma] {plot_csvs/ensemble_multimodal_predscatter.csv};
    \end{axis}
    \end{tikzpicture} &
    \begin{tikzpicture}
    \begin{axis}[
        width=0.24\paperwidth,
        colormap name=viridis,
        view={0}{90},
    ]
    \addplot3 [surf,shader=interp] table [x=x, y=y, z=z, col sep=comma] {plot_csvs/ensemble_epistemic_contours.csv};
    \addplot3 [contour gnuplot={contour dir=z, draw color=black, number=10, labels=false}, opacity=0.2, shader=interp]
        table [x=x, y=y, z=z, col sep=comma] {plot_csvs/ensemble_epistemic_contours.csv};
    \addplot [only marks, mark=*, color=deep2]
        table [x=test_x, y=test_y, col sep=comma] {plot_csvs/ensemble_epistemic_predscatter.csv};
    \addplot [only marks, mark=*, color=deep1]
        table [x=test_x, y=pred, col sep=comma] {plot_csvs/ensemble_epistemic_predscatter.csv};
    \end{axis}
    \end{tikzpicture}  \\
    \makebox[5ex]{\raisebox{12ex}{\rotatebox[origin=c]{90}{\bf Energy model}}} &
    \begin{tikzpicture}
    \begin{axis}[
        width=0.24\paperwidth,
        xlabel={x},
        ylabel={y},
        colormap name=viridis,
        view={0}{90},
    ]
    \addplot3 [surf,shader=interp] table [x=x, y=y, z=z, col sep=comma] {plot_csvs/ebr_homosced_contour.csv};
    \addplot3 [contour gnuplot={contour dir=z, draw color=black, number=10, labels=false}, opacity=0.2, shader=interp]
        table [x=x, y=y, z=z, col sep=comma] {plot_csvs/ebr_homosced_contour.csv};
    \addplot [only marks, mark=*, color=deep2]
        table [x=test_x, y=test_y, col sep=comma] {plot_csvs/ebr_homosced_predscatter.csv};
    \addplot [only marks, mark=*, color=deep1]
        table [x=test_x, y=pred, col sep=comma] {plot_csvs/ebr_homosced_predscatter.csv};
    \end{axis}
    \end{tikzpicture} &
    \begin{tikzpicture}
    \begin{axis}[
        width=0.24\paperwidth,
        xlabel={x},
        colormap name=viridis,
        view={0}{90},
    ]
    \addplot3 [surf,shader=interp] table [x=x, y=y, z=z, col sep=comma] {plot_csvs/ebr_heterosced_contour.csv};
    \addplot3 [contour gnuplot={contour dir=z, draw color=black, number=10, labels=false}, opacity=0.2, shader=interp]
        table [x=x, y=y, z=z, col sep=comma] {plot_csvs/ebr_heterosced_contour.csv};
    \addplot [only marks, mark=*, color=deep2]
        table [x=test_x, y=test_y, col sep=comma] {plot_csvs/ebr_heterosced_predscatter.csv};
    \addplot [only marks, mark=*, color=deep1]
        table [x=test_x, y=pred, col sep=comma] {plot_csvs/ebr_heterosced_predscatter.csv};
    \end{axis}
    \end{tikzpicture} &
    \begin{tikzpicture}
    \begin{axis}[
        width=0.24\paperwidth,
        xlabel={x},
        colormap name=viridis,
        view={0}{90},
    ]
    \addplot3 [surf,shader=interp] table [x=x, y=y, z=z, col sep=comma] {plot_csvs/ebr_multimodal_contour.csv};
    \addplot3 [contour gnuplot={contour dir=z, draw color=black, number=10, labels=false}, opacity=0.2, shader=interp]
        table [x=x, y=y, z=z, col sep=comma] {plot_csvs/ebr_multimodal_contour.csv};
    \addplot [only marks, mark=*, color=deep2]
        table [x=test_x, y=test_y, col sep=comma] {plot_csvs/ebr_multimodal_predscatter.csv};
    \addplot [only marks, mark=*, color=deep1]
        table [x=test_x, y=pred, col sep=comma] {plot_csvs/ebr_multimodal_predscatter.csv};
    \end{axis}
    \end{tikzpicture} &
    \begin{tikzpicture}
    \begin{axis}[
        width=0.24\paperwidth,
        xlabel={x},
        colormap name=viridis,
        view={0}{90},
    ]
    \addplot3 [surf,shader=interp] table [x=x, y=y, z=z, col sep=comma] {plot_csvs/ebr_epistemic_contour.csv};
    \addplot3 [contour gnuplot={contour dir=z, draw color=black, number=10, labels=false}, opacity=0.2, shader=interp]
        table [x=x, y=y, z=z, col sep=comma] {plot_csvs/ebr_epistemic_contour.csv};
    \addplot [only marks, mark=*, color=deep2]
        table [x=test_x, y=test_y, col sep=comma] {plot_csvs/ebr_epistemic_predscatter.csv};
    \addplot [only marks, mark=*, color=deep1]
        table [x=test_x, y=pred, col sep=comma] {plot_csvs/ebr_epistemic_predscatter.csv};
    \end{axis}
    \end{tikzpicture} \\
    \end{tabular}
    \vspace{-2mm}
    \caption{
        Visualization of the predicted density on the test set for trained models.
        Contour plots: Log-likelihood output from each model.
        Yellow: the high-density region;
        Blue: the low-density region.
        Blue points: Predicted mean.
        Orange points: Test set.
    }
    \vspace{-2mm}
    \label{fig:contour_plots}
\end{figure*}
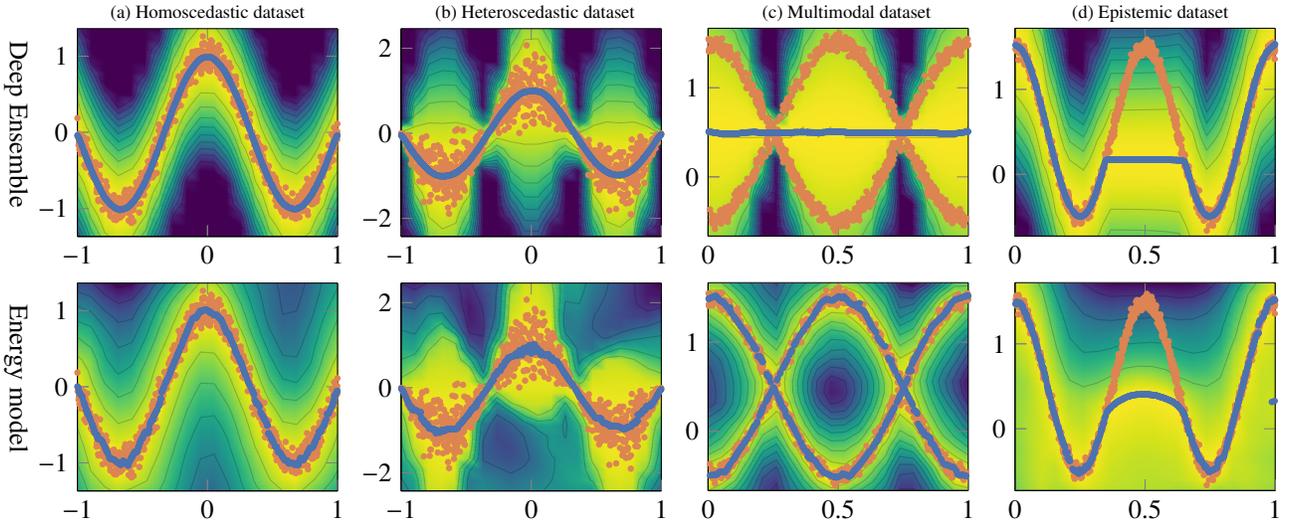

For each synthetic dataset, we train one EBR model, and one DE model.
Both types of models were trained with a batch size of 128 using the Adam \citep{kingma15} optimizer.
The sizes of the training, validation, and testing datasets were respectively,
100000, 1000, and 1000 samples.
Fig. \ref{fig:contour_plots} shows the resulting log-likelihood functions from the models,
along with their predictions on our test sets.

\parsection{Energy Based Regression}
We implement EBR as a 9-layer perceptron with ReLU activations and a hidden size of 256,
which we train for 20 epochs with a fixed learning rate $10^{-4}$.
While 9 layers may sound too deep for our simple toy datasets, our initial experiments showed clear signs of underfitting with shallower models.

\parsection{Deep Ensemble}
All models in our ensemble are identical 5-layer perceptrons, with ReLU activations, and a hidden size of 256.
In this case 5 layers is more than enough for the model to learn our toy datasets,
and we choose this size simply to avoid a comparison between models of drastically
different depth.
Overfitting is not a concern since we can control the amount of data we generate. 
We train five models for 20 epochs each with a fixed learning rate of $10^{-3}$.
In our experiments we use the predicted variance as $U(x)$, more choices are compared by \citet{xiong2023hinge}.

\subsection{Stability on varying test set sizes}


All metrics are estimates that are computed on a test set of finite size. It is thus important to investigate how quickly each metric converges to its expected value, and whether the estimates are biased for small test set sizes.

\begin{figure}[ht!]
    \centering
    \begin{tikzpicture}
    \begin{axis}[
        width=0.3\paperwidth,
        xmode=log,
        xtick=data,
        xticklabels={{}, $2^4$, {}, $2^6$, {}, $2^8$, {}, $2^{10}$, {}, $2^{12}$, {}, $2^{14}$, {}, $2^{16}$},
        xlabel={},
        ylabel={Metric value},
        title={\bf (a) Iterative data collection},
        title style={
            rotate=90,
            anchor=south,
            at={(-0.4,0.45)},
        },
        legend style={
            nodes={scale=0.75, transform shape},
            at={(0.2,0.5)},
            anchor=west,
            legend columns=2,
            /tikz/every even column/.append style={column sep=0.1cm}
        },
    ]
    \addplot table [x=test_size, y=ause, col sep=comma] {plot_csvs/subset_size_experiment.csv};
    \addplot table [x=test_size, y=spearman, col sep=comma] {plot_csvs/subset_size_experiment.csv};
    \addplot table [x=test_size, y=nll, col sep=comma] {plot_csvs/subset_size_experiment.csv};
    \addplot table [x=test_size, y=ece, col sep=comma] {plot_csvs/subset_size_experiment.csv};
    \legend {
        AUSE,
        Spearman,
        NLL,
        CE,
    }
    \end{axis}
    \end{tikzpicture}
    \begin{tikzpicture}
    \begin{axis}[
        width=0.3\paperwidth,
        xmode=log,
        xtick=data,
        xticklabels={{}, $2^4$, {}, $2^6$, {}, $2^8$, {}, $2^{10}$, {}, $2^{12}$, {}, $2^{14}$, {}, $2^{16}$},
        xlabel={\footnotesize Dataset size},
        ylabel={Metric value},
        title style={
            rotate=90,
            anchor=south,
            at={(-0.4,0.4)},
        },
        title={\bf (b) Mean of 100 i.i.d. datasets},
        legend style={
            nodes={scale=0.75, transform shape},
            at={(0.2,0.5)},
            anchor=west,
            legend columns=2,
            /tikz/every even column/.append style={column sep=0.1cm}
        },
    ]
    \addplot table [x=test_size, y=mean_ause, col sep=comma] {plot_csvs/test_size_experiment.csv};
    \addplot table [x=test_size, y=mean_spearman, col sep=comma] {plot_csvs/test_size_experiment.csv};
    \addplot table [x=test_size, y=mean_nll, col sep=comma] {plot_csvs/test_size_experiment.csv};
    \addplot table [x=test_size, y=mean_ece, col sep=comma] {plot_csvs/test_size_experiment.csv};
    \legend{
        AUSE,
        Spearman,
        NLL,
        CE,
    }
    \end{axis}
    \end{tikzpicture}
    \vspace{-2ex}
    \caption{
        Experiments to test two types of stability of metrics under different test dataset sizes.
    }
    \label{fig:stability}
    \vspace{-2ex}
\end{figure}
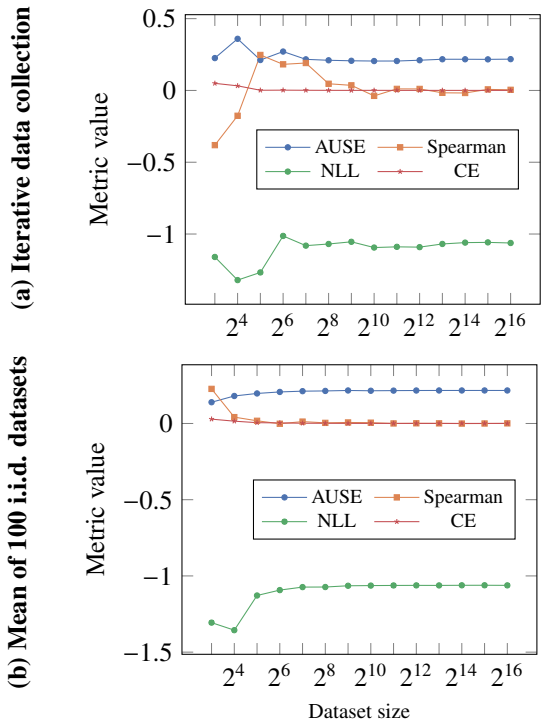

First, in order to test convergence,
we emulate a process of iteratively collecting points.
We begin with generating a single large test dataset of size $2^{16}$.
From this dataset, we randomly sample data points without replacement,
which allows us to investigate how quickly each metric converges to its \say{true} value for a given dataset.
We evaluate each metric at test set sizes $2^3$, $2^4$, \ldots, $2^{16}$.
The resulting metrics from each subset are reported in Fig. \ref{fig:stability} (a).


Second, in order to investigate any bias in the approximation of each metric,
we sample 100 i.i.d. datasets at each size $2^3$, $2^4$, \ldots, $2^{16}$.
We then report the average of each metric over these 100 datasets in Fig. \ref{fig:stability} (b).
If a metric is truly unbiased, the average at a small dataset size should converge to the same value as the average at larger dataset sizes.

For both experiments, we use our heteroscedastic dataset because we believe
it is the best candidate to highlight any stability issues, due to its varying noise levels.
We apply the DE model in both experiments since its performance is similar to EBR model on the heteroscedastic dataset as shown in Fig. \ref{fig:contour_plots} (b) and the choice of model is agnostic.

\subsection{Metrics under different types of uncertainty}
We compute all metrics for both our models on the test sets of each of the four datasets.
As a reference, we compare the results to those computed on the data-generating distribution.
Using the data generating distribution will, in expectation, minimize NLL and CE,
and should provide a good point of comparison for all metrics.
We summarize metrics of both models for each dataset in Tab. \ref{tab:homo}, \ref{tab:hetero}, \ref{tab:multimodal} and \ref{tab:epistemic} respectively.

\begin{table}[ht!]
    \vspace{-2ex}
    \centering
    \scriptsize
    \caption{Metrics for all models on Homoscedastic dataset.}
    \begin{tabular}{l p{12ex} p{12ex} p{12ex} p{12ex}}
        \toprule
        & \textbf{AUSE}$\downarrow$ & \textbf{CE}$\downarrow$ & \textbf{NLL}$\downarrow$ & \textbf{Spearman}$\uparrow$ \\
        \hline
        & \\[-2ex] 
        \textbf{DE}       & 0.5915 & 0.0023 & -0.8819 & 0.0444 \\
        \textbf{EBR}   & 0.5707 & 0.0032 & -0.8568 & -0.0264 \\
        \textbf{True dist.}     & 0.5917 & 0.0003 & -0.8965 & 0.0076 \\
         \bottomrule
    \end{tabular}
    \label{tab:homo}
    \vspace{-2.5ex}
\end{table}

\begin{table}[ht!]
     \vspace{-2.5ex}
    \centering
    \scriptsize
    \caption{Metrics for both models on Heteroscedastic dataset.}
     \begin{tabular}{l p{12ex} p{12ex} p{12ex} p{12ex}}
        \toprule
        & \textbf{AUSE}$\downarrow$ & \textbf{CE}$\downarrow$ & \textbf{NLL}$\downarrow$ & \textbf{Spearman}$\uparrow$ \\
        \hline
        & \\[-2ex] 
        \textbf{DE}       & 0.2334 & 0.0001 & -0.1091 & 0.0005 \\
        \textbf{EBR}   & 0.2422 & 0.0001 & -0.0990 & -0.0419 \\
        \textbf{True dist.}     & 0.2305 & 0.0001 & -0.1472 & 0.0280 \\
         \bottomrule
    \end{tabular}
    \label{tab:hetero}
     \vspace{-2ex}
\end{table}

\begin{table}[ht!]
    \vspace{-2.5ex}
    \centering
    \scriptsize
    \caption{Metrics for both models on multimodal dataset.}
     \begin{tabular}{l p{12ex} p{12ex} p{12ex} p{12ex}}
        \toprule
        & \textbf{AUSE}$\downarrow$ & \textbf{CE}$\downarrow$ & \textbf{NLL}$\downarrow$ & \textbf{Spearman}$\uparrow$ \\
        \hline
        & \\[-2ex] 
        \textbf{DE}       & 0.0071 & 0.0229 & 0.8098  & -0.0420  \\
        \textbf{EBR}   & 0.5821 & 0.0018 & -0.6535 & 0.0093 \\
        \textbf{True dist.}     & 0.5180 & 0.0001 & -0.7935  & 0.0073 \\
         \bottomrule
    \end{tabular}
    \label{tab:multimodal}
     \vspace{-2.5ex}
\end{table}

\begin{table}[ht!]
    \vspace{-2.5ex}
    \centering
    \scriptsize
    \caption{Metrics for both models on epistemic dataset.}
     \begin{tabular}{l p{12ex} p{12ex} p{12ex} p{12ex}}
        \toprule
        & \textbf{AUSE}$\downarrow$ & \textbf{CE}$\downarrow$ & \textbf{NLL}$\downarrow$ & \textbf{Spearman}$\uparrow$ \\
        \hline
        & \\[-2ex] 
        \textbf{DE}       & 0.6016 & 0.0145 & 36.4332 & 0.0298 \\
        \textbf{EBR}   & 1.3888 & 0.0298 & 31.2425 & 0.0067 \\
        \textbf{True dist.}     & 0.5454 & 0.0001 & -1.5871 & -0.0194 \\
         \bottomrule
    \end{tabular}
    \label{tab:epistemic}
    \vspace{-2ex}
\end{table}

\subsection{Metrics for Real-world Applications}
To further demonstrate the behaviour of the metrics, we apply them to the real-world problem of stereo disparity prediction. 
This is a regression task where the disparity of X-coordinates are to be predicted for corresponding pixels in the left and right frames of a stereo camera.  
We adopt the continuous disparity network(CDN) from \cite{garg2020wasserstein},
and we use the Hinge-Wasserstein training loss \citep{xiong2023hinge} (hinge value 0.005) as it improves both the regression and UQ performance.
The model is trained on the large dataset Sceneflow (\cite{mayer2016large}), where dense ground truth disparity maps are provided.
As this is real image data, there is already an unknown amount of aleatoric uncertainty in the test set. To this we add controllable amounts of uniform Gaussian noise, to both left and right frames during testing.
Specifically, Gaussian noise with a set of different variances is added to the RGB image normalized by ImageNet mean and variance. (This means that the image variance can be expected to be around 1 in the noise free case.) 
We use End-Point-Error (EPE) and 1-Pixel Threshold Error (1PE), as in \cite{garg2020wasserstein} to quantify for regression performance.
All relevant performance and uncertainty metrics are provided in Tab.\ \ref{tab:real}.


\begin{table*}[!ht]
    \centering
    \caption{All metrics for stereo disparity prediction with CDN supervised by hinge-Wasserstein (hinge value 0.005). Variances are reported with 5 runs.}
    \label{tab:real}
    \begin{tabular}{cllllll}
    \hline
    Noise scale& EPE    & 1px       & NLL       & AUSE  & CE      & Spearman \\
0   & 1.01 {\footnotesize$\pm    0$}    & 9.44   {\footnotesize$\pm    0$}    & 0.000521{\footnotesize$\pm3.6\times10^{-6}$}  & 0.159 {\footnotesize$\pm 0.0006$} & 0.2663 {\footnotesize$ \pm 0    $  }& 0.0012{\footnotesize$ \pm 0.0004$ }\\ \hline
0.1 & 1.13 {\footnotesize$\pm 0.00$}    & 10.97  {\footnotesize$\pm 0.00$}    & 0.000590{\footnotesize$\pm4.2\times10^{-6}$}  & 0.161 {\footnotesize$\pm 0.0014$} & 0.2629 {\footnotesize$ \pm 0.0001$ }& 0.0010{\footnotesize$ \pm 0.0006$ }\\ \hline
0.2 & 1.85 {\footnotesize$\pm 0.00$}    & 15.91  {\footnotesize$\pm 0.00$}    & 0.000735{\footnotesize$\pm2.8\times10^{-6}$}  & 0.128 {\footnotesize$\pm 0.0013$} & 0.2457 {\footnotesize$\pm 0       $}& 0.0006{\footnotesize $\pm 0.0006 $}\\ \hline
0.3 & 2.81 {\footnotesize$\pm 0.01$}    & 22.54  {\footnotesize$\pm 0.01$}    & 0.000904{\footnotesize$\pm1.2\times10^{-6}$}  & 0.113 {\footnotesize$\pm 0.0008$} & 0.2210 {\footnotesize$\pm 0       $}& 0.0024{\footnotesize $\pm 0.0005 $}\\ \hline
0.4 & 4.01 {\footnotesize$\pm 0.01$}    & 29.81  {\footnotesize$\pm 0.01$}    & 0.001061{\footnotesize$\pm1.9\times10^{-6}$}  & 0.103 {\footnotesize$\pm 0.0008$} & 0.1947 {\footnotesize$\pm 0.0002  $}& 0.0026{\footnotesize $\pm 0.0006 $}\\ \hline
0.5 & 5.50 {\footnotesize$\pm 0.01$}    & 37.32  {\footnotesize$\pm 0.01$}    & 0.001212{\footnotesize$\pm2.0\times10^{-6}$}  & 0.099 {\footnotesize$\pm 0.0008$} & 0.1686 {\footnotesize$ \pm 0.0001$ }& 0.0040{\footnotesize$ \pm 0.0757$ }\\ \hline
    \end{tabular}
\end{table*}

\section{Discussion}
Here we analyze the experiment results, and reason about various strengths and limitations of each metric.
\subsection{Stability on varying test set sizes}
\label{sec:stability}
From Fig. \ref{fig:stability} (a) we can see that all metrics vary as the test set sizes increase and finally converge to the expectation. 
We can conclude that all metrics converge beyond a dataset size of $2^{10}$. 
Hence, for most modern regression applications,
where test datasets are larger than $\sim$1000 samples,
stability should not be a cause for concern.
While our analysis is based only on a single dataset,
we expect the convergence speed to be reasonably consistent across different datasets.
Though the amplitudes of metrics vary from each other,
arguably we can also conclude that CE is
the most stable and AUSE converges second fastest.

From Fig. \ref{fig:stability} (b) we can see that the averages change for different dataset sizes.
We will henceforth refer to this behavior as \textit{estimation bias of mean}.
Arguably, estimation bias is undesirable,
since we expect the average of a metric to converge to the same value regardless of the dataset size.
As such, a metric with significant bias of the mean should be considered unstable and its use discouraged.
Overall, CE is the most stable and AUSE the second most in terms of bias of the mean.
We can also conclude that no metric seems to exhibit any meaningful bias beyond a dataset size of $2^6$.

Together, these two plots
suggest that the most stable metric is CE, followed by AUSE, then NLL, and finally Spearman correlation.
Surprisingly, in both Fig. \ref{fig:stability} (a) and (b), Spearman correlation has large (absolute) values with small test set sizes and gradually converges to zero as the test set size grows.
This can be explained by a closer look at Fig.~\ref{fig:contour_plots} (b): 
All points close to $x=0$ will have similar predicted uncertainty values,
but their errors may be vastly different,
which counteracts the correlation between error and uncertainty.
Naturally, this counteracting effect becomes less impactful when samples are sparse,
and Spearman correlation has been used successfully on for example the rMD17 dataset
\citep{christensen2020rMD17}.

\subsection{Metrics under different types of uncertainty}

\parsection{Homoscedastic and Heteroscedastic datasets:}
\label{sec:homo&hetero}
In Fig. \ref{fig:contour_plots} (a) and (b) we can see that both models learn predictive distributions close to the generating functions (defined in Sec.\ \ref{sec:datasets}).

In Tab.~\ref{tab:homo} and \ref{tab:hetero},
all metrics except Spearman suggest that both models learn the homoscedastic and heteroscedastic distributions approximately equally well.
\begin{figure}[t!]
\vspace{-1ex}
    \centering
    \begin{tikzpicture}
    \begin{groupplot}[
        group style={group size=2 by 1},
        width=0.235\paperwidth,
        xlabel={$\alpha$},
        ylabel={Relative MAE},
        legend style={
            nodes={scale=0.75, transform shape},
            at={(1.2,-0.4)},
            anchor=north,
            legend columns=3,
            rotate=180,
            /tikz/every even column/.append style={column sep=0.2cm}
        },
        xmin=0,
        xmax=1,
        xtick={0, 1},
        xtick align=outside,
        xtick pos=lower,
        ymin=0,
        ytick align=outside,
        ytick pos=left,
        ytick={0, 1},
        no markers
    ]
    \nextgroupplot[
        title={\scriptsize{(a) Homoscadestic}},
        title style={
            at={(0.5, 0.9)},
            anchor=south
        }
    ]
    \addplot table [x=fraction, y=oracle, col sep=comma] {plot_csvs/ensemble_homo_ause.csv};
    \addplot table [x=fraction, y=sparsification, col sep=comma] {plot_csvs/ensemble_homo_ause.csv};
    \addplot table [x=fraction, y=sparsification, col sep=comma] {plot_csvs/homosced_ause_true_std.csv};
    \legend{
        Oracle,
        Ensemble,
        True Distribution
    }
    \nextgroupplot[
        title={\scriptsize{(b) Heteroscadestic}},
        title style={
            at={(0.5, 0.9)},
            anchor=south
        },
        ylabel={}
    ]
    \addplot table [x=fraction, y=oracle, col sep=comma] {plot_csvs/ensemble_hetero_ause.csv};
    \addplot table [x=fraction, y=sparsification, col sep=comma] {plot_csvs/ensemble_hetero_ause.csv};
    \addplot table [x=fraction, y=sparsification, col sep=comma] {plot_csvs/heterosced_ause_true_std.csv};
    \end{groupplot}
    \end{tikzpicture}
    \vspace{-2ex}
    \caption{
        Sparsification plot from Deep Ensemble and True Distribution for the homoscedastic and heteroscedastic datasets. $\alpha$ is the fraction of removed samples. 
    }
    \label{fig:sparsification}
    \vspace{-2ex}
\end{figure}
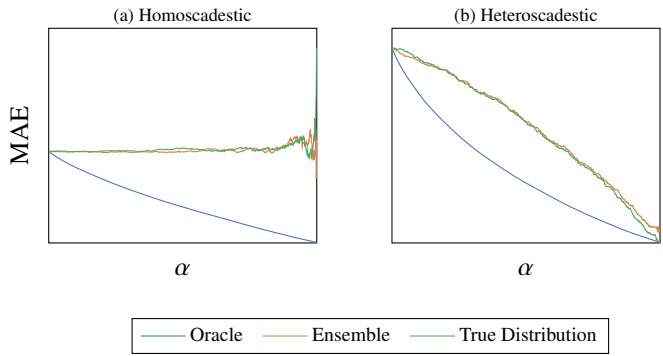
In Fig.~\ref{fig:sparsification} (a), the sparsification curve is nearly horizontal, indicating that it is impossible to learn a meaningful correlation between errors and uncertainty.
The predicted variance can only partially explain the error, the remaining part is caused by the difference between the annotation and the predicted mean. Intuitively, we expect this scenario of a perfectly uniform variance to be rare in real-world applications.
This also indicates that generating uncertainty measures from the predicted parametric distribution may not be the optimal approach for sorting samples. 
Fig.~\ref{fig:contour_plots} (a) and Fig.~\ref{fig:sparsification} (b) together further show this: even when the predictive and generating distributions are very well aligned, the AUSE value is not 0.
For the heteroscedastic dataset, the true distribution achieves the best NLL, and the DE model is slightly better than EBR model.
This is because the DE model learns a sharper distribution than EBR, as can be seen in the contour plots of Fig.~\ref{fig:contour_plots} (a).
NLL, as a proper scoring rule, addresses calibration and sharpness simultaneously \citep{gneiting2007strictly}. 

\parsection{Multimodal dataset:}
In Fig.~\ref{fig:contour_plots} (c) we can see that energy-based model successfully learns to predict both modes, but as the ensemble is not designed to output multimodal predictions, it predicts the average of the modes, and a wide distribution.

Tab.~\ref{tab:multimodal} shows larger discrepancy between metrics.
All metrics suggest that the energy-based model has learned a good distribution that is close to the generating distribution.
AUSE for DE is near prefect, which indicates good detection of when DE fails to predict multiple modes.
CE and NLL instead indicate the ensemble's failure to learn both modes with high values.
This highlights the main difference between AUSE (and Spearman correlation) and CE/NLL:
AUSE measures a correlation between uncertainty and the absolute error,
while CE and NLL measure how well the learned distribution captures the data.
This raises the question of which type of measurement is most useful, which we will discuss in Sec.\ \ref{sec:conclusion}.

\parsection{Epistemic dataset:}
In Fig. \ref{fig:contour_plots} (d), both models exhibit somewhat larger uncertainty in the OOD region.
However, neither model predicts a distribution that assigns high probability to the test data.
Visually, the predicted distributions of both models look similar.
The only obvious perceptible difference is the slight bend in the center of the distribution of the EBR model.

In Tab.~\ref{tab:epistemic}, AUSE indicates that the ensemble predicts distribution that can find OOD samples almost as well as the generating distribution, while the distribution from the EBR model is much worse.
Both CE and NLL agree that both models failed to learn a good distribution,
but they disagree on which model is worse.
The contours in  Fig.~\ref{fig:contour_plots} (d) also shows the DE model gives sharper distribution.
It is worth noting that there is an entire field of study of OOD detection, which offers better methods for handling epistemic uncertainty.
Here, we are instead interested in whether or not each metric can detect a failure to learn the underlying distribution.

Looking at the AUSE scores of 0.6016 and 1.3888 for the ensemble and EBR models respectively,
we can infer that this small bend reduces the correlation between error and uncertainty enough to (more than) double the AUSE score.
Intuitively, we would expect these two similar distributions to get similar scores, which is exactly the case with both CE and NLL.
This raises questions regarding the stability of AUSE under small model variations.

\subsection{Real-world Applications}
Tab.\ \ref{tab:real} summarize the results of the real-world stereo disparity experiment. As can be expected, regression performances drop as the noise scale increases. 
It is immediately clear that the variance of Spearman correlation is too large to be a reliable metric.
Both AUSE and CE keeps decreasing as the noise scale increases, which indicates that the used Hinge--Wasserstein loss is able to quantify uncertainty under adversarial noise.
Meanwhile, NLL keeps increasing, which means the sharpness of predictions get worse with higher noise levels. This is consistent with \cite{xiong2023hinge} which illustrates that a lack of sharpness is a side effect of the Hinge--Wasserstein loss.

\subsection{Interpretability}
CE has well defined bounds which makes it easy to interpret.
With NLL and AUSE, this is more difficult. We can compare them across models, however, the lower and upper bounds depend on both model and data.

\parsection{CE:}  CE has a lower bound 0 when the predicted and generating distributions are equal.
The upper bound, however, is dependent on the number of intervals and the weights $w_j$.

\parsection{NLL:}
The lower bound of NLL is given by the generating distribution, which is typically unknown when training for real-world deployment.
The upper bound is positive infinity.

\parsection{AUSE:}
From Sec.~\ref{sec:homo&hetero} we know that the generating distribution may not even give the optimal lower bound 0.
We argue that predicted variance can only partially explain the error.
There is no clear upper bound for AUSE. 

In summary, NLL and AUSE can only tell us which model is better at UQ but cannot tell us how good it is.
This lack of interpretability constitutes a severe limitation of applying AUSE and NLL into real-world autonomous systems.

\subsection{Concluding Remarks}
\label{sec:conclusion}

We have tested four common uncertainty assessment metrics on four synthetic datasets that are designed to illustrate properties of these metrics. 
We also test all the metrics on a real-world regression task, Stereo disparity.
Below we summarize our conclusions:

\parsection{Spearman Correlation} This metric is in general unsuitable. It has stability issues for small test sets, and converges to zero as the test set size grows, see Sec.~\ref{sec:stability}.

\parsection{AUSE} 
 This metric offers a robust quantized measure on the correlation between predictive uncertainty and errors compared with Pearson correlation and Spearman correlation. It tells us how wrong a model prediction is likely to be, which is vital for trust-worthy autonomous systems. On the down side, its value is unbounded and thus it lacks interpretability, and it may fail in tasks where homoscedastic uncertainty dominates (but this is rare in real-world applications).
 It may be more appropriate for loss prediction \citep{8954021, cui2024learning}, which is another UQ approach compared to predicting full distributions.

\parsection{Calibration Error}
This metric has a clear lower bound of 0, and is highly interpretable; It requires the least amount of samples to be stable, 
and can be extended to general confidence intervals \cite{kuleshov2018}. Limitations are that it requires predicting a distribution, and it does not capture how wrong the predictions can be.

\parsection{NLL} This measures the shape of the predictive distributions, including calibration and sharpness.
It fails when the generating distribution is multimodal but the training target is unimodal \citep{xiong2023hinge}, and it requires the largest amount of test samples to be stable among those tested here.

Essentially CE and NLL measure how well the predicted probability distribution corresponds to the true data generating distribution, but CE is the more interpretable of the two.
In contrast, AUSE and Spearman measure how well the predicted uncertainty can tell the magnitude of errors.
We discourage the use of Spearman correlation for uncertainty quantification, as AUSE has consistently proved to be more robust.
Which metric is most useful will naturally differ between applications.
\parsection{Acknowledgement}: 
This work was funded by Swedish national strategic research environment ELLIIT, grant C08. Computations for this work were enabled by the National Academic Infrastructure for Supercomputing in Sweden (NAISS) and the Berzerlius resource provided by National Supercomputing Centre at Linköping University and the Knut and Alice Wallenberg foundation, partially funded by the Swedish Research Council through grant agreement no. 2022-06725. This project is also supported by the Wallenberg AI, Autonomous Systems, and Software Program (WASP) funded by the Knut and Alice Wallenberg Foundation.
\vspace{-2ex}

\bibliographystyle{model5-names}
\bibliography{references}

\end{document}